# Deepfake Detection Analyzing Hybrid Dataset Utilizing CNN and SVM

JACOB MALLET

University of Wisconsin-Eau Claire

LAURA PRYOR

University of Wisconsin-Eau Claire

DR. RUSHIT DAVE

Minnesota State University, Mankato

DR. MOUNIKA VANAMALA

University of Wisconsin-Eau Claire

*Abstract*— Social media is currently being used by many individuals online as a major source of information. However, not all information shared online is true, even photos and videos can be doctored. Deepfakes have recently risen with the rise of technological advancement and have allowed nefarious online users to replace one's face with a computer-generated face of anyone they would like, including important political and cultural figures. Deepfakes are now a tool to be able to spread mass misinformation. There is now an immense need to create models that are able to detect deepfakes and keep them from being spread as seemingly real images or videos. In this paper, we propose a new deepfake detection schema using two popular machine learning algorithms; support vector machine and convolutional neural network, along with a publicly available dataset named the 140k Real and Fake Faces to accurately detect deepfakes in images with accuracy rates reaching as high as 88.33%.

**CCS CONCEPTS** • Computing Methodologies ~ Machine Learning ~ Machine Learning Approaches • Security and Privacy ~ Human and Societal Aspects of Security and Privacy • Computing Methodologies ~ Artificial Intelligence ~ Computer Vision ~ Computer Vision Problems ~ Object Detection

## 1 INTRODUCTION

In the current state of the world, what important figures say and do matters greatly. However, with newly created technology, anyone is now able to easily impersonate anyone they would like, even important cultural and political figures, using deepfakes. A deepfake, while it can be either a photo or a video, allows for a nefarious actor to replace his or her face with a computer-generated version of an authentic person's face.

The use of deepfakes has allowed many nefarious people to correctly fool others across the globe into thinking that the video or photo they made posing as the person of interest (POI) is a real image or video [1-3]. As the dependency on social media grows, so does the ability for false information to be spread and believed by the people who consume it. If a video or photo is created well enough that the average user is unable to spot the difference, anything said or done in that piece of media can be consumed by these average users and considered true information. Thus, there has become a large need for software to be developed to accurately detect deepfakes in order to stop the potential threat of widespread misinformation.

This paper proposes a new deepfake detection scheme for images using two different machine learning algorithms to evaluate its performance. As seen in related security fields [4-12], the use of Machine Learning is continuing to advance and has been able to perform effectively in its given scheme. Machine Learning's success in other security-related frameworks leads us to believe that it is the perfect tool to be used in software that detects deepfakes. The two machine learning algorithms tested are Support Vector Machine (SVM) and Convolutional Neural Networks (CNN). Both models have been found to perform well in deepfake detection frameworks [13-16]. Additionally, our model uses data from a Kaggle dataset that complied 70,000 images from the Flickr-Faces-HQ Dataset [17] and 70,000 images generated by StyleGAN, which has been widely used to generate deepfakes in previous research [18-20]. Our model developed in this paper will use two machine learning algorithms, SVM and CNN, to evaluate its ability to correctly recognize deepfake images.

The most common tactic to create deepfakes is to use generative adversarial networks (GAN) which generate fake images or videos of the POI. The main goal in deepfake detection is for the detection model to discover the minute inconsistencies between the image depicting the POI and the GAN-generated face. The basis of this study was built on previous research in [21], in which we found the use of temporal features to be the most consistent deepfake detection feature set regarding effectiveness. Additionally, we found CNN and SVM to be two well-performing machine learning algorithms used with deepfake detection schemas.

## 2 RELATED WORK

### 2.1 SVM Classification

Similar to our study, researchers in [22] also use SVM as a classifier in their model. [22] works with videos rather than pictures. Initially, the videos are minimized to 20 frames with the idea of reducing the computational cost of the model. For every frame, OpenFace2, developed in [], was used to extract the face from the frame by identifying facial landmarks. [22] turns to other research, taken from and utilizes ten pre-trained CNN models for feature extraction. The reasoning behind this is that these models have been trained on massive databases and possess the ability to quickly and accurately identify important keypoints. This is due to these pre-trained CNN models having already identified and computed sample keypoints meaning they do not have to be learned again, helping with the computation costs associated with this model. What the models will be learning is the important features of the new images, such as face texture, eyes, nose orientation, lip size, and more. After these features are extracted, [22] chooses to classify with SVM based on the algorithm's robustness and the ability it possesses to deal with over-fitted training data. Accuracies ranged from 89% to 98% for all ten of the different pre-trained CNN models. This model shows the potential of using SVM to classify deepfakes.

## 2.2 CNN Classification

[23] aims to classify deepfakes using a CNN model that utilizes multiple different streams to perform classification via ensemble voting. [23] evaluates their model on multiple datasets, including the one used in this study, the 140k Real and Fake Faces dataset. Three CNN models are developed, with two being some of the same pre-trained models seen in the previous study. The CNN model developed in [23] deploys a 2D convolutional layer followed by a max pooling layer to reduce the image size. Lastly a dense output layer with a sigmoid activation function used for each model to classify the images is added. After each model classifies the image, the voting stage is initiated. Two voting approaches are used in [23], soft voting ensemble and hard voting ensemble. Hard voting ensemble sums the votes from all the models and the image is classified based on the majority of the votes, while soft voting ensemble sums the predicted probabilities from all the models and classifies the image based on the class with the largest sum probability. The models were evaluated without voting and with. Accuracies reached as high as 95.97% from their CNN model on the 140k Real and Fake Faces dataset, and as low as 52.37% from the pre-trained VGG16 model. Accuracies increased dramatically when all three models were combined to classify by ensemble voting. All 3 models scored in the 50% to 60% accuracy range for the Real and Fake Face dataset, but when ensemble voting was applied an accuracy of 75.13% was achieved. Results improved across the board, with an accuracy as high as 98.79% from the 140k Real and Fake Faces dataset. [23] displays a unique voting system while using multiple models to achieve high accuracy scores.

Researchers in [24] employ three different CNNs to classify images from the DFDC and DeepFakeTIMIT dataset. Both of these datasets are composed of videos, but the datasets are preprocessed to only include one face frame from each video. This resulted in 270,000 images for training only from the DFDC dataset. [24] extracted the faces from the images using a MTCNN library. Several image augmentations, including image cropping, random horizontal flips, gaussian blur, gaussian noise, and more were applied to the dataset in order to avoid overfitting. Three different pre-trained CNN classifiers were used in [24], with them training the last several layers of the different classifiers. Each classifier makes a prediction individually, and the average is taken from the three numbers to classify the frame as real or fake. One limitation that plagues many models is the lack of transferability. [24] claims that models typically perform well on the dataset they're trained on, but when applied to other datasets the accuracy takes a sizable dropoff. This model aims to test this by strictly training on the DFDC dataset, but testing on the DeepFakeTIMIT, where the model achieves a 96.50% accuracy. This accuracy score shows that the fusion of the three pre-trained models improved the results than any of the classifiers on their own. Interestingly, [24] also finds that the performance of CNN improves when testing low resolution images rather than high resolution. This model displays the capabilities of using CNN as a classifier, as well as highlighting the transferability of CNN with datasets other than the training dataset.

## 3 METHODOLOGY

### 3.1 System Overview

In this research, we test two different classification algorithms, CNN and SVM. We will overview both models in this section.

*3.1.1 SVM Classification Model*

For the first model, we use a CNN initially, followed by a SVM to classify the images. The first layer of the model is a 2D convolutional layer. This layer contains 32 filters to learn from. Padding is set to remain the same in order to preserve spatial dimensions of the volume. The stride is set to 2, which means the convolutional layer will be traversing the image by stepping 2 pixels at a time. A 2D max pooling layer follows the 2D convolutional layer. This layer is utilized to reduce the computational cost of traversing the image and effectively reduce the size. These two layers are again added to the CNN model, leaving the model with four layers thus far. In order for SVM to be applied effectively, a flatten layer is added to make the input data one dimensional. Finally, a dense output layer is added with a linear activation function and a kernel regularizer from keras called l2. This regularizer is used to reduce some of the less significant features identified by the model. The model is compiled with a hinge loss function using the adam optimizer.

*3.1.2 CNN Classification Model*

The second model applied in this study performs CNN classification. The first 2 layers of the model are 2 2D convolutional layers with the same parameters as the SVM model. Following the convolutional layers is the first 2D max pooling layer, which again effectively reduces the image size by taking the maximum value outputted by the 2D convolutional layers over a 2 by 2-pixel size filter. To avoid overfitting, we add a dropout layer with a rate of 0.2. This layer is followed by a flatten layer and fed to a fully connected dense layer. Another identical dropout layer is added to prevent overfitting again before the last layer. Finally, the output layer is added with a sigmoid activation function. This activation function will output a value between 0 and 1, which represents the probability of the image belonging to the positive class, or being a real image, according to our model. The model's output is rounded up or down to 0 or 1 for binary classification. Compilation is done with an Adam optimizer and a binary cross-entropy loss.

**3.2 Dataset**

The dataset used in this study is called 140k real and fake faces [25]. This dataset consists of 70,000 deepfakes taken from a million deepfakes generated by a style GAN. 70,000 real faces from the Flickr-Faces-HQ dataset join the deepfakes in the dataset, creating an equal amount of deepfakes and real images. This is essential so our model has equal exposure to real and fake faces. Data augmentation is applied to the dataset used our model to artificially increase the size, with random horizontal flips, shearing, and zooming being applied.

**3.3 Ethical Disclaimer**

Not only does the use of deepfakes have safety concerns regarding misinformation, but they also come with a multitude of ethical issues as well. The use of deepfakes, whether nefarious or not, uses a person's face without permission unless the face is completely computer generated. It is important to recognize that in our evaluation of our methodology, we are using a dataset filled with completely computer-generated faces. Thus, none of the synthetic faces were created with the intent of taking a real person's face and generating it on someone else. Additionally, we also include a dataset of real faces that were collected with consent from the users. At no time throughout this study was there the use of images of either real or synthetic faces without the consent of the participants that allowed for their likeness to be used in the collection of the original datasets.

**3.4 Classifier Background**

This section will discuss the background of the two classifiers, namely SVM and CNN, that were utilized in our model.

*3.4.1 SVM*

SVM is a supervised machine learning algorithm used quite frequently in deepfake detection schemas using temporal features. The basis of SVM is that it uses hyperplanes to accurately create predictions. In SVM, multiple hyperplanes are held in an n-dimensional space in which n is the number of features processed in the model. Additionally, in this n-dimensional space, there are data points called support vectors. Support vectors are the furthest data point out from a distinct group of data. From these hyperplanes and support vectors, the algorithm's goal is to find a single hyperplane that has the most amount of distance between each group's support vector. This hyperplane acts as a decision boundary for classification. SVM wants to find the hyperplane with the maximum amount of distance between each group because the larger the distance between each group, the better likelihood that each data point will be very distinct to that specific group. Thus, the goal is to find a hyperplane that has a large margin between the different classification groups to ensure confident predictions.

*3.4.2 CNN*

CNN is a deep learning algorithm that has frequently been used in deepfake detection models due to its high performance when dealing with images and videos. CNNs are designed, in a way, to allow computers to mimic the way that human brains can process and distinguish different objects. For this process to be done, CNNs need to take the 3-dimensional image and simplify it into a 1-dimensional representation without losing any important aspects of the image that can be used for classification. CNNs are comprised of three main layers: convolution, pooling, and fully-connected layers. In the convolutional layer, a weighted filter is used in multiple, in most cases, 3x3 areas of the image to calculate a dot-product for that area. This filter shifts throughout the entire image until the once 2-dimensional image is now a sequence of dot-products called a feature map. The main purpose of the pooling layer is to reduce the size of the feature map to reduce computational needs and the chance of overfitting. Finally, the fully-connected layer is a multi-layer perceptron that is used to classify the image. Each of these layers can be repeated multiple times in a single CNN, with multiple of a single layer type in a row, however, the overall order of each layer must always be the same.

**3.5 Success Metrics**

In order to evaluate the correctness of our model we used six success metrics: accuracy (ACC), precision, recall, F1 Score, Area Under the Curve (AUC) and a confusion matrix. The most used success metric, ACC, measures the percentage of how many images our model correctly classified in relation to the total amount of images. A confusion matrix outputs the number of classifications that fall into each one of the following categories: true positive, true negative, false positive, and false negative. A true positive in the case of this model would be an image of a real person being classified as a real image, whereas a true negative is an image of a synthetic person that was classified as a synthetic image. False negatives are real images that are classified as synthetic images and false positives are synthetic images that are classified as real. Precision looks to measure the exactness of the model; therefore, it looks at the percentage of true positive classifications out of all classifications that were predicted as positive. Precision is often measured along with recall which examines the sensitivity of a model. Recall is the measurement of the percentage of true positive classifications out of all classifications that were real positives. F1 Score is another calculation of accuracy, but instead of looking at all possible classifications, it only uses precision and recall so it only looks at accuracy using positive classifications. Finally, AUC is the area under the receiver operating characteristics (ROC) curve. The ROC curve uses the rate of true positives and the rate of false positives to create a curve of probability. The area under that curve, AUC, then shows how efficient our model is in distinguishing deepfakes as deepfakes and real images as real images. Thus, the larger the AUC the better our model is in distinguishing the difference between the two different classes.

Table 1: Comparison of Algorithms

| Success Metrics | CNN | SVM |
| --- | --- | --- |
| Accuracy | 88.33% | 81.69% |
| Precision | 89.91% | 84.82% |
| Recall | 86.34% | 77.19% |
| AUC | 88.33% | 81.69% |
| F1 Score | 88.10% | 80.82% |

## 4 RESULTS & DISCUSSION

The most significant finding from our results is that CNN produced the best results for all success metrics tested. While the two algorithms used to evaluate the performance of our model performed similarly to one another, CNN ended up performing slightly better across all success metrics used. Using CNN resulted in a top accuracy of 88.33% and SVM obtained an accuracy of 81.69%. Additionally, our model achieved precisions of 89.93% and 84.82% for CNN and SVM respectively. Thus, our model performed well in its ability to correctly classify real images as real and synthetic images as fake. Sensitivity performed similarly to the exactness as CNN obtained an 86.34% recall and SVM a 77.19% recall. Having high precision and recall results are good indicators of the usability of our model. A high recall indicates that our model is less likely to reject a real image and high precision indicates that our model is trustworthy in its ability to correctly identify real and synthetic images. However, the slightly lower recalls in comparison to precision could also indicate that our model is slightly too insensitive and is allowing too many false negatives. This makes sense when we look at our confusion matrix, as we can see for both algorithms that when our model is making incorrect identifications, most of those identifications are classifying a synthetic image as a real image. Our confusion matrices can be seen in Tables 2 and 3. When creating a curve using our true and false positives, we were able to achieve an AUC of 88.33% for CNN and an AUC of 81.69% for SVM. This means that when running CNN, our model has just over an 88% chance of correctly distinguishing between the two classes and when running SVM, almost an 82% chance of correctly distinguishing. These high AUCs show that our model can effectively distinguish between a deepfake and a real image. Finally, when it came to the F1 Score, both models performed well, but again the model using CNN achieved a higher score of 88.10% compared to SVM's 80.82%.

Table 2: Confusion Matrix for CNN

|  | Real | Deepfake |
|---|---|---|
| Real | 15808 | 1692 |
| Deepfake | 2391 | 15109 |

## 5 LIMITATIONS

The urgency for accurate deepfake detection models only grows as an increasing amount of news is spread online. One of the most common limitations that plagues many studies across deepfake detection research, including this one, is the transferability of the model. Applying a model that was trained on a singular dataset to the real world or on other datasets, oftentimes the performance drops dramatically. Piecing together a training dataset that includes deepfakes from diverse sources would be something to explore in future research. Another limitation that is present in this study is the cost of the model to run in a real-world scenario. The only way to detect deepfakes is to test all of the images, or at least the videos or images that are identified as communicating important information. Running these models at that scale would be extraordinarily expensive considering the vast amount of images and videos uploaded to the internet every second.

Table 3: Confusion Matrix for SVM

|  | Real | Deepfake |
|---|---|---|
| Real | 15082 | 2418 |
| Deepfake | 3992 | 13508 |

## 6 CONCLUSION & FUTURE WORK

In this paper, we introduced a new deepfake detection model focused on the detection of deepfakes in images. In this model, we used a publicly available dataset that combined real images from the FFHQ dataset and computer-generated images generated using styleGAN. Additionally, we tested the performance of our model using two different machine learning algorithms: Support Vector Machine and Convolutional Neural Networks. From this model we were able to achieve accuracies as high as 88.33% using our Convolutional Neural Networks classifier.

While our model was able to achieve promising results in its ability to accurately detect deepfake images, there is still more that needs to be done for this model to be applicable to real-world instances. As mentioned previously, when a model that was only trained on a single dataset is used in real world applications, its performance often reduces immensely. More work needs to be done to increase our model's ability to be transferred into real-world settings. This can be done by improving our model by training and testing it on more robust datasets to be able to limit the decrease in performance as much as possible. Additionally, we look to test this model with other machine learning algorithms to ensure that we equip our model with the best performing algorithms currently available.

## ACKNOWLEDGMENTS

Funding for this project has been provided by the University of Wisconsin-Eau Claire's Karlgaard Computer Science Scholarship Foundation.

DOI:http://dx.doi.org/10.1109/cvprw56347.2022.00016

[21] Jacob Mallet, Rushit Dave, Naeem Seliya, and Mounika Vanamala. 2022. Using deep learning to detect deepfakes, *9th International Conference on Soft Computing & Machine Intelligence (ISCMI)* (2022).

[22] Momina Masood, Marriam Nawaz, Ali Javed, Tahira Nazir, Awais Mehmood, and Rabbia Mahum. 2021. Classification of deepfake videos using pre-trained convolutional Neural Networks. *2021 International Conference on Digital Futures and Transformative Technologies (ICoDT2)* (2021). DOI:http://dx.doi.org/10.1109/icodt252288.2021.9441519

[23] Jatin Sharma, Sahil Sharma, Vijay Kumar, Hany S. Hussein, and Hammam Alshazly. 2022. Deepfakes Classification of Faces Using Convolutional Neural Networks. *Traitement du Signal* 39, 3 (2022), 1027–1037. DOI:http://dx.doi.org/10.18280/ts.390330

[24] Sohail Ahmed Khan, Hang Dai, and Alessandro Artusi. 2021. Adversarially robust deepfake media detection using fused convolutional neural network predictions. *CoRR* abs/2102.05950 (2021). DOI:http://dx.doi.org/doi.org/10.48550/arXiv.2102.05950

[25] Xhlulu. 2020. 140k real and fake faces. Retrieved December 15, 2022 from https://www.kaggle.com/datasets/xhlulu/140k-real-and-fake-faces
9